\documentclass[conference]{IEEEtran}
\usepackage[left=0.635in,right=0.635in,top=0.75in,bottom=1.05in]{geometry}
\usepackage{amsmath,amsfonts}
\usepackage{algorithmic}
\usepackage{algorithm}
\usepackage{array}
\usepackage[caption=false,font=normalsize,labelfont=sf,textfont=sf]{subfig}
\usepackage{textcomp}
\usepackage{stfloats}
\usepackage{url}
\usepackage{verbatim}
\usepackage{graphicx}
\usepackage{cite}
\usepackage{hyperref}
\usepackage{verbatim}
\usepackage{listings}
\usepackage{xcolor}
\usepackage{booktabs}

\lstdefinestyle{jsonlike}{
    basicstyle=\ttfamily\small,
    showstringspaces=false,
    breaklines=true,
    frame=single,
    backgroundcolor=\color{gray!10},
    stringstyle=\color{green!50!black},
    keywordstyle=\color{blue!80!black}\bfseries,
    morekeywords={true,false,null,"mission_type","location","road","from","to",
                  "objectives","sensors","processing_tasks","output","format",
                  "content","priority"}
}
\begin{document}

\title{AI and Semantic Communication for Infrastructure Monitoring in 6G-Driven Drone Swarms}

\author{
    \IEEEauthorblockN{Tasnim Ahmed, 
    Salimur Choudhury}
    
    \IEEEauthorblockA{School of Computing,
    Queen's University, Ontario, Canada}

    \IEEEauthorblockA{\{tasnim.ahmed, s.choudhury\}@queensu.ca}
}

\maketitle

\begin{abstract}
The adoption of unmanned aerial vehicles to monitor critical infrastructure is gaining momentum in various industrial domains. Organizational imperatives drive this progression to minimize expenses, accelerate processes, and mitigate hazards faced by inspection personnel. However, traditional infrastructure monitoring systems face critical bottlenecks---5G networks lack the latency and reliability for large-scale drone coordination, while manual inspections remain costly and slow. We propose a 6G-enabled drone swarm system that integrates ultra-reliable, low-latency communications, edge AI, and semantic communication to automate inspections. By adopting LLMs for structured output and report generation, our framework is hypothesized to reduce inspection costs and improve fault detection speed compared to existing methods.
\end{abstract}

\begin{IEEEkeywords}
6G Networks, Drone Swarm, Infrastructure Management, Semantic Communication, Edge Computing, Energy Efficiency.
\end{IEEEkeywords}

\section{Introduction}
Drones, key components of the unmanned aerial system, have been widely used over the past decade for various applications, including inspections, surveillance, delivery, search and rescue, etc. \cite{app13031256}. Drones equipped with sensing capabilities are particularly well-suited for inspections in critical environments that are challenging or inaccessible for humans. Recently, there has been growing interest in AI-equipped drones, which not only perform inspections but also make decisions on the fly. However, operating AI models onboard significantly impacts flight times, as their intelligent inspection capabilities are constrained by limited flight duration and energy resources. Consequently, for autonomous inspections over moderately large geographical areas, the deployment of a cooperative swarm of drones becomes essential. While collaborating through the transmission of raw sensor data among drones is possible, it remains highly resource-intensive, limiting the computing capabilities and processor performance of the drones.

With advancements in intelligent communication and various technologies, the applications of Semantic Communication (SC) are increasingly being explored across diverse fields such as Unmanned Aerial Vehicle (UAV) communications, intelligent transportation, and healthcare \cite{LIU2024528}. SC enhances machine-to-machine and machine-to-human communication by enabling intelligent exchanges. Semantic extraction technologies are being developed to improve communication efficiency. For instance, Chakravarthy et al. \cite{9689963} propose a deep learning approach that processes UAV vision sensor data in real-time, segmenting captured scenes with robust semantics. In this method, UAVs with limited battery capacity focus on data collection, while computationally intensive tasks are offloaded to cloud servers.

Building on these advancements, onboard AI capabilities can significantly aid in scene recognition and fault detection across various infrastructures.
Infrastructure monitoring has historically relied on ground-based visual inspections. To reduce labour costs, literature has introduced various autonomous platforms, such as unmanned ground vehicles and drones. Several studies have explored the use of drones for inspecting power lines, bridges, and railways \cite{app13031256}, etc., employing sensor, image, and multimodal data. For example, Jalil et al. \cite{s19133014} utilized infrared and RGB images to detect faults in power lines. However, most research focuses on the effectiveness of a single aerial vehicle, despite the limitation of their flight duration. Additionally, current studies typically apply these systems to inspect only one type of infrastructure at a time, leaving a gap in research regarding intelligent inspection of multiple infrastructures or addressing multiple objectives simultaneously. While SC holds great potential for facilitating an automated suite for infrastructure inspection, complete management still requires human expertise to give instruction or produce professional reports from the gathered semantic data. Manual approaches often lead to inefficiencies and a lack of comprehensive assessments. Given the success of Large Language Models (LLMs) in automating report generation in fields like healthcare \cite{WANG2023100033} and construction \cite{PU2024124601}, we hypothesize that state-of-the-art LLMs can generate professional-grade reports from semantic data collected by drone swarms, creating a fully automated system for infrastructure inspection and management. We propose that this system can be conceptualized as an Internet of Everything (IoE) system, which extends the concept of the Internet of Things (IoT) by integrating people, processes, data, and things into a unified network. The evolution of wireless networks has traditionally been driven by the need for higher data rates, but the rise of IoE systems, connecting millions of people and billions of devices, signals a shift from rate-centric enhanced mobile broadband services toward ultra-reliable, low-latency communications (URLLC). While 5G networks support basic IoE and URLLC services, their capacity to sustain the unprecedented growth of new IoE applications is debatable \cite{8869705}. To meet the demands of such IoE services, which require simultaneous delivery of high reliability, low latency, and high data rates across diverse devices, a 6G wireless system is necessary.

To this end, we propose an end-to-end system where a swarm of AI-equipped drones autonomously manages infrastructure inspection and fault detection tasks across various domains, including buildings, roads, and electrical equipment. Central to our approach is a local LLM-based server that processes user queries, generates structured outputs, and orchestrates the swarm operations. Each drone, equipped with onboard AI and sensors, follows the LLM instructions to execute specific tasks while engaging in semantic communication with other drones to optimize flight paths and share processed data. Upon completion, the drones transmit all gathered insights back to the LLM for comprehensive report generation that ensures efficient and accurate infrastructure management. The key contributions of this research can be summarized as follows---

\begin{enumerate}
    \item We present an automated system for infrastructure inspection that utilizes the power of 6G, semantic communication, edge AI, and LLMs to generate professional-grade reports from drone-collected semantic data.
    \item We highlight the necessary drivers, requirements, and enabling technologies of the envisioned 6G wireless systems for this framework.
    \item We show a high-level use case of the proposed system to demonstrate its effectiveness in real-world scenarios.
\end{enumerate}

\section{System Model}
\begin{figure}[htp]
\centering
\includegraphics[width=\columnwidth]{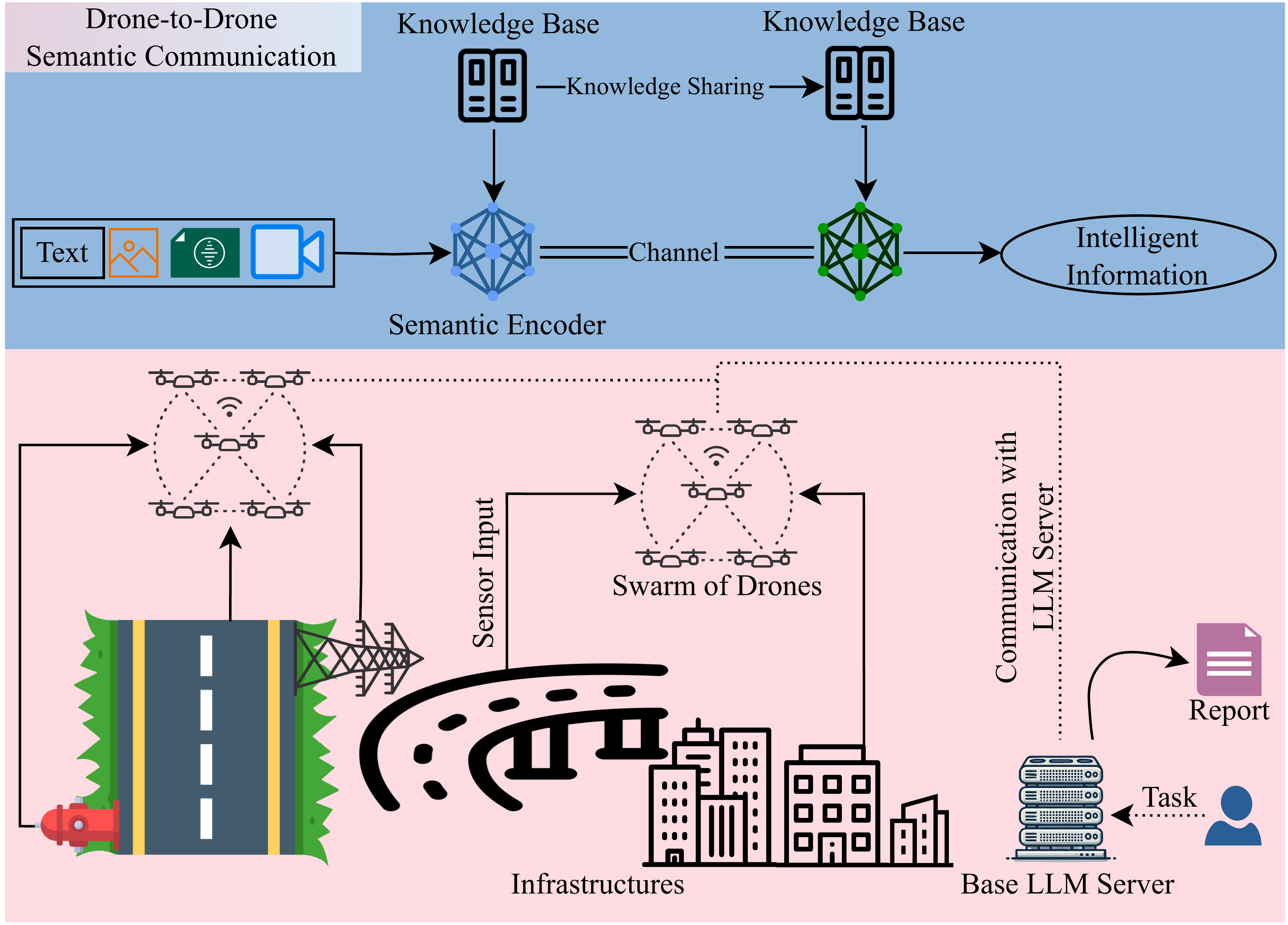}
\caption{\textbf{Proposed system model.} LLM processes unstructured user requests in natural language, and corresponding tasks are assigned to the drones. The swarm of drones collects multimodal data from the environment and processes it with onboard AI. SC facilitates the transmission of processed information (shown in the upper part of the figure). The combined information is passed to the LLM server for report generation.}
\label{fig:framework}
\end{figure}

The proposed system model is depicted in \figureautorefname~\ref{fig:framework}. The system model is designed to monitor a diverse range of critical infrastructures, including buildings, roads, bridges, electric poles, fire hydrants, roads, etc. The swarm of drones, equipped with advanced sensors, is capable of performing tasks such as routine inspections, fault detection, damage severity assessment, and construction progress monitoring. SC plays a crucial role in this system by enabling drones to exchange semantically rich data, significantly enhancing energy efficiency by reducing the need to transmit large volumes of raw sensor data. Each drone captures multimodal sensor data and encodes it using a semantic encoder, typically powered by an AI-based model. The encoded data is then transmitted wirelessly to the target drone, which decodes it using a semantic decoder. Both the encoder and decoder rely on a knowledge base, which can be either private or shared among the drones and may vary depending on the specific task at hand. It allows drones to intelligently distribute tasks, such as identifying inspection targets and prioritizing faults, thereby optimizing their collective efforts while conserving energy.

\subsection{Concept of Operation}
A monitoring mission begins with the user providing unstructured input and describing the task in natural language. The base LLM server transforms this input into a data structure that defines the mission objectives and includes relevant details such as the mission type, perimeter, and sensor attributes. Each mission is constrained by the available energy of the drones involved. Using the structured output from the LLM, a cooperative path planner calculates the energy required for data collection and onboard AI processing. The planner dynamically allocates tasks based on real-time battery status, e.g., drones with less than 30\% battery prioritize low-power roles such as semantic relaying or communication bridging and are rerouted to charging stations, while those with sufficient reserves handle compute-intensive tasks like LiDAR mapping or thermal imaging. We hypothesize that this energy-aware strategy extends swarm operational time significantly compared to static task allocation. Based on the location and limitations of the drone, the planner determines the flight path for each drone, including charging stations, frequencies, and specific tasks for each drone in the swarm. Once this is defined, the mission paths and tasks are assigned to the corresponding drone swarms.

\begin{figure*}[htp]
\centering
\includegraphics[width=0.8\textwidth]{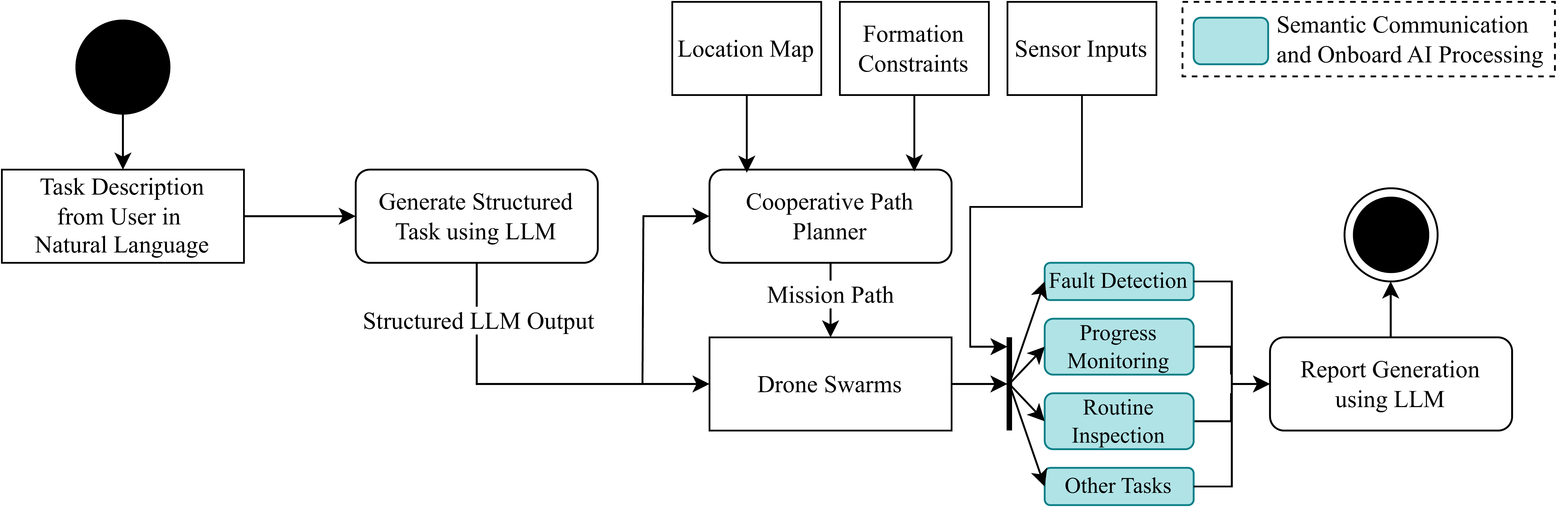}
\caption{Activity diagram for infrastructure monitoring.}
\label{fig:activity}
\end{figure*}

The drones execute their assigned tasks through SC. This intelligent communication allows drones within a swarm to efficiently distribute tasks. For instance, some drones may handle communication with the base LLM server for transmitting semantic data, others may focus on data collection, and others may perform onboard AI processing tasks like object detection, instance segmentation, image and video processing, and noise reduction. In cases of fault or anomaly detection, the onboard processing can assess whether additional inspections are needed for the detected issue. Drones can also adjust their flight paths based on processed semantic data. The processed data from each swarm is then sent back to the base LLM server, which generates a report tailored to the specific task and user needs. These reports are valuable for both expert and non-expert personnel, as they help extract meaningful information and patterns from the raw sensor data. Typically, obtaining such insights would require extensive post-processing, a manual, labour-intensive, and time-consuming process. The LLM-generated reports optimize this process and make it easier to interpret the drone outputs. A critical aspect of this operation is URLLC, which is envisioned to be supported by 6G technology. Without URLLC, coordination among drones and real-time monitoring of various infrastructures would be impossible. An activity diagram illustrating this operation is provided in \figureautorefname~\ref{fig:activity}.  

\section{Foundations of 6G for Intelligent Infrastructure Monitoring}
\subsection{Drivers}
The evolution towards 6G is driven by the exponential growth in mobile traffic, the demand for URLLC, and the need for ubiquitous connectivity. As mobile devices and smart infrastructures increase, the demand for high-capacity, low-latency networks becomes increasingly critical \cite{singh20246gevolutionenhancementsinnovations}. In particular, the rising complexity and scale of connected devices, such as autonomous drones, necessitate a network that can support seamless, real-time communication across vast areas, ensuring reliable and efficient operations.

\subsection{Enabling Technologies \& Alignment with 6G Objectives}
To meet these demands, 6G networks will rely on several advanced technologies. Terahertz communication and millimetre-wave technology will be pivotal in providing the necessary bandwidth and data rates \cite{8732419}. Additionally, the integration of AI and machine learning at the network edge will enhance the intelligence and autonomy of the system, allowing for more efficient task distribution and energy management among drone swarms \cite{8808168}.

The proposed framework aligns closely with the emerging technical objectives outlined in 6G standardization efforts, particularly in addressing the limitations of 5G networks for large-scale autonomous systems. As highlighted in recent 6G roadmaps, next-generation networks prioritize URLLC with end-to-end delays below $1$ ms and packet loss rates under \(10^{-7}\) to support mission-critical applications such as drone swarms \cite{8869705}. Our system directly leverages these capabilities to enable real-time coordination between drones during infrastructure inspections, where delays exceeding this limit could lead to collisions or missed fault detections. This also aligns with the vision of 6G to support tactile internet applications, which demands deterministic latency guarantees for synchronized operations on distributed devices \cite{9145564}. Furthermore, our integration of SC addresses the emphasis of 6G on energy-efficient data transmission. Traditional drone swarms transmit raw sensor data, consuming up to 85\% of their energy budget on communication. By contrast, our SC framework is projected to reduce data payloads by at least 60–70\% through context-aware encoding as demonstrated with preliminary experiments. This efficiency gain supports the goal of 6G to achieve an improvement of $100\times$ in energy efficiency over 5G systems. The shared knowledge base facilitates semantic encoding by ensuring that distributed edge nodes keep their context models aligned, thereby reducing redundant data transmissions. The reliance on edge AI also reflects the shift of 6G towards AI-native air interfaces, where neural networks are deeply embedded in the communication stack. Finally, the use of LLMs for automated reporting aligns with the intent to integrate generative AI into network management planes, enabling human-centric service customization---a key pillar in the 6G roadmap for industrial automation.

\subsection{Requirements}
For the proposed infrastructure monitoring system to function effectively, 6G must fulfill several key requirements. First, it must provide ultra-reliable, low-latency communication to ensure that data from drones is transmitted and processed in real-time to provide immediate responses to detected faults or anomalies. Second, the network must support massive machine-type communications to simultaneously handle the vast number of connected devices, such as drones and sensors. Finally, 6G must offer enhanced energy efficiency and spectrum efficiency to prolong the operational life of the drones and reduce the overall power consumption of the system \cite{8869705}.

\section{Preliminary Implementation}
As a proof of concept, we demonstrate a high-level implementation of the proposed system. Our preliminary implementation includes converting user prompts to structured LLM output and semantically encoding sensor data (image).

\subsection{Hardware and Software Design}
\begin{figure}[htbp]
\centering
\includegraphics[width=0.65\columnwidth]{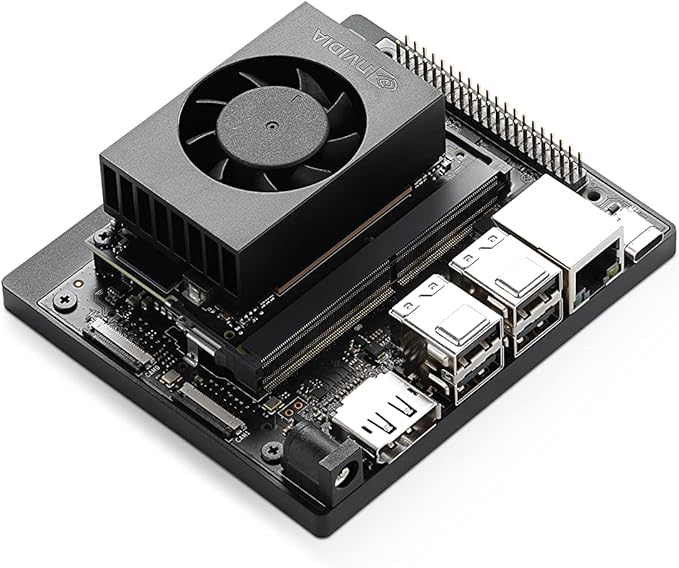}
\caption{Embedded device for onboard AI processing on drones.}
\label{fig:jetson}
\end{figure}

We used a mainstream computer as our base LLM server, equipped with an Intel Core™ i$9$-$14900$KF processor with $24$ cores, $68$ MB cache, and a maximum clock speed of $6.0$ GHz. An NVIDIA GeForce RTX $4090$ GPU with $24$ GB GDDR6X memory handled the computational tasks. The system was supported by $64$ GB of RAM, and the Llama-$3.1$-$8$B served as the backbone LLM for our initial evaluation. For onboard AI and semantic encoding on the drones, we relied on the Nvidia Jetson Orin Nano (refer to \figureautorefname~\ref{fig:jetson}), a low-power embedded system that operates within a $7$W-$15$W range and weighs less than $200$ grams, allowing it to function autonomously even in remote locations. This device has a 6-core Arm® Cortex®-A$78$AE $64$-bit CPU with $1.5$ MB L$2$ and $4$ MB L$3$ cache, and a $625$ MHz $1024$-core Nvidia Ampere architecture GPU with $32$ Tensor Cores, supported by $8$ GB $128$-bit LPDDR$5$ memory, capable of performing up to $40$ TOPS. The entire edge AI system can be powered by a $15$W DC power source.

\subsection{Structured LLM Output}
\begin{figure*}[htbp]
\centering
\includegraphics[width=\textwidth]{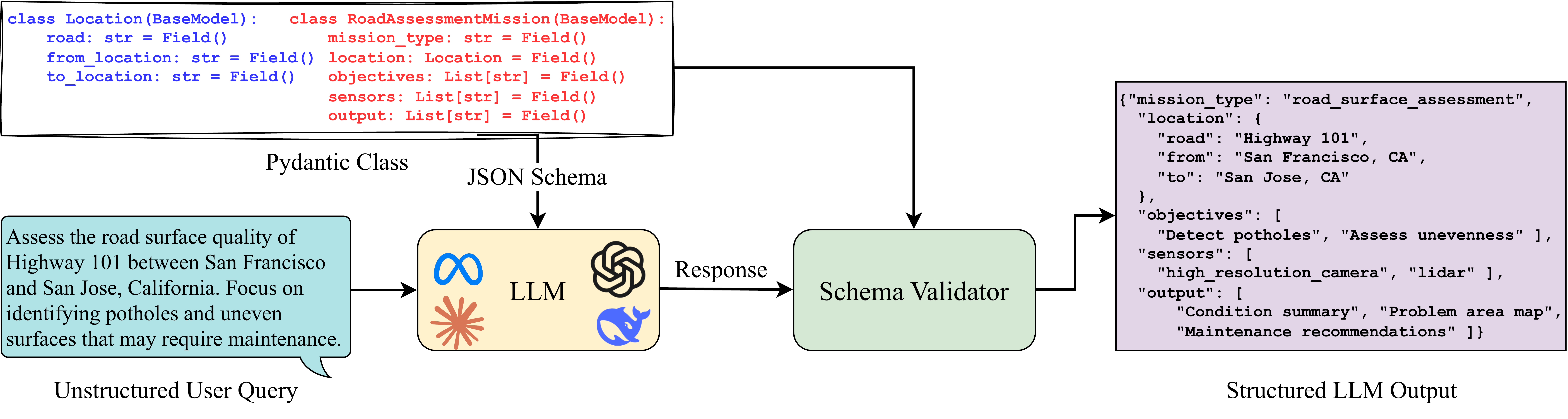}
\caption{Structured Output from LLM with Pydantic Class}
\label{fig:structured}
\end{figure*}

One of the critical aspects of our proposed pipeline is the requirement for fully automated instruction generation for drones, implying that the determination of the path planning should proceed without external annotations or human involvement, simply from unstructured human command/query.
To automate this task, we must process user queries in a structured format.
Our initial tests with open-source LLMs indicate that the LLM responses do not consistently follow a single pattern across different types of user queries---even when the LLM sampling temperature is set to a very low value.
This presents an intriguing paradox: LLMs inherently generate unstructured text, reflecting their training on massive, mostly unstructured datasets.
To achieve formatted responses, we propose a structured output-parsing approach, where each LLM inference is accompanied by a predefined data model (class definition) that helps extract the structured command. 
\figureautorefname~\ref{fig:structured} illustrates our structured data model for road inspection tasks, which captures various aspects like mission objectives, sensor requirements, and expected outputs. This structured approach is essential not only for the inspection itself but also for automated report generation and analysis. The figure also demonstrates how we transform unstructured user requests into well-defined, structured formats that can effectively coordinate drone swarms, trigger the right edge AI models, and produce comprehensive reports. Furthermore, by implementing structured LLM output schemas in SC systems, we can achieve significant bandwidth reduction through the elimination of redundant information and standardization of message formats.

\subsection{Onboard AI Processing}
\begin{figure}[htbp]
\centering
\includegraphics[width=0.8\columnwidth]{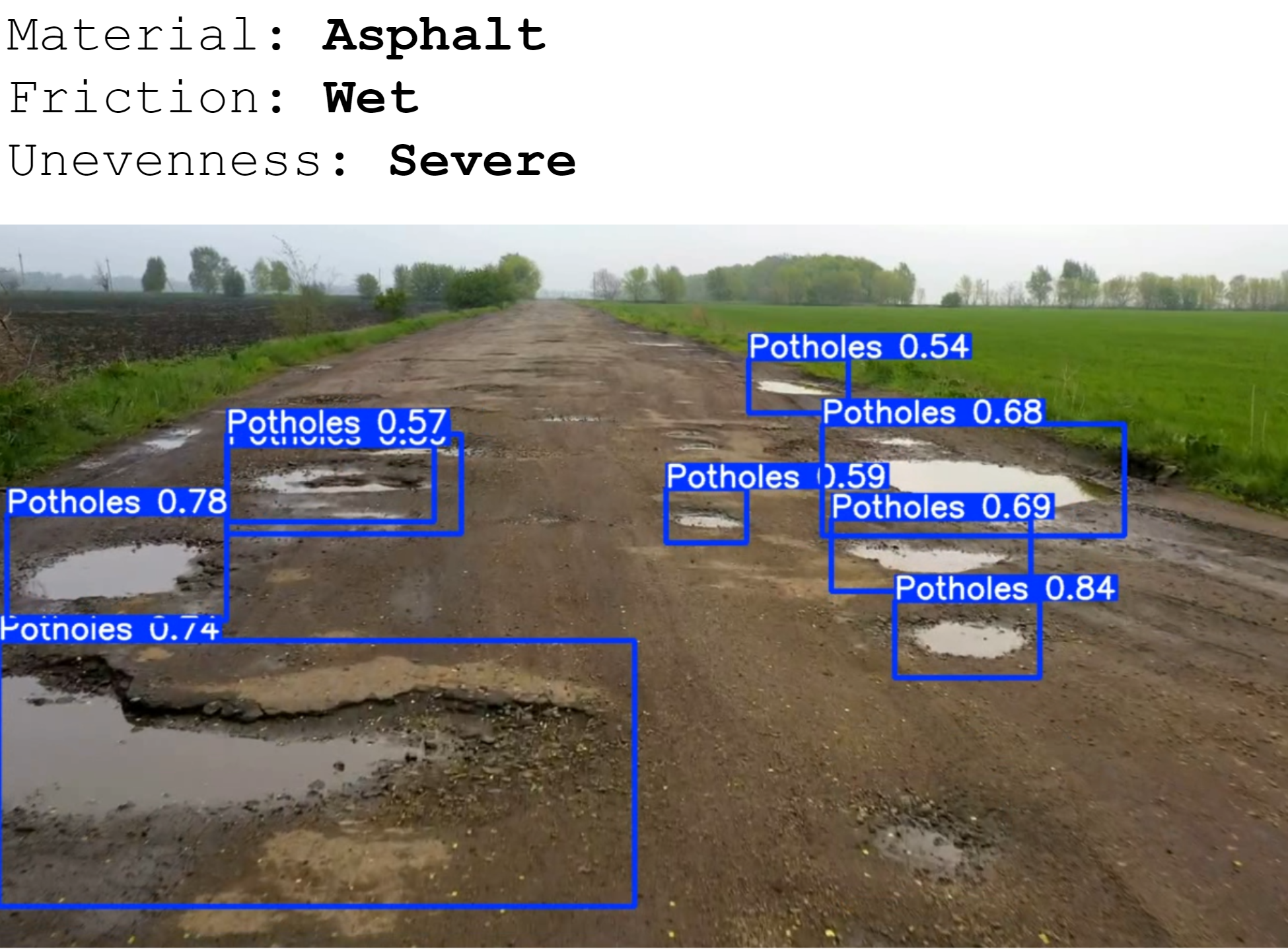}
\caption{Road condition monitoring using edge AI device from video frames.}
\label{fig:jetson_output}
\end{figure}

\begin{figure}[htbp]
\centering
\includegraphics[width=\columnwidth]{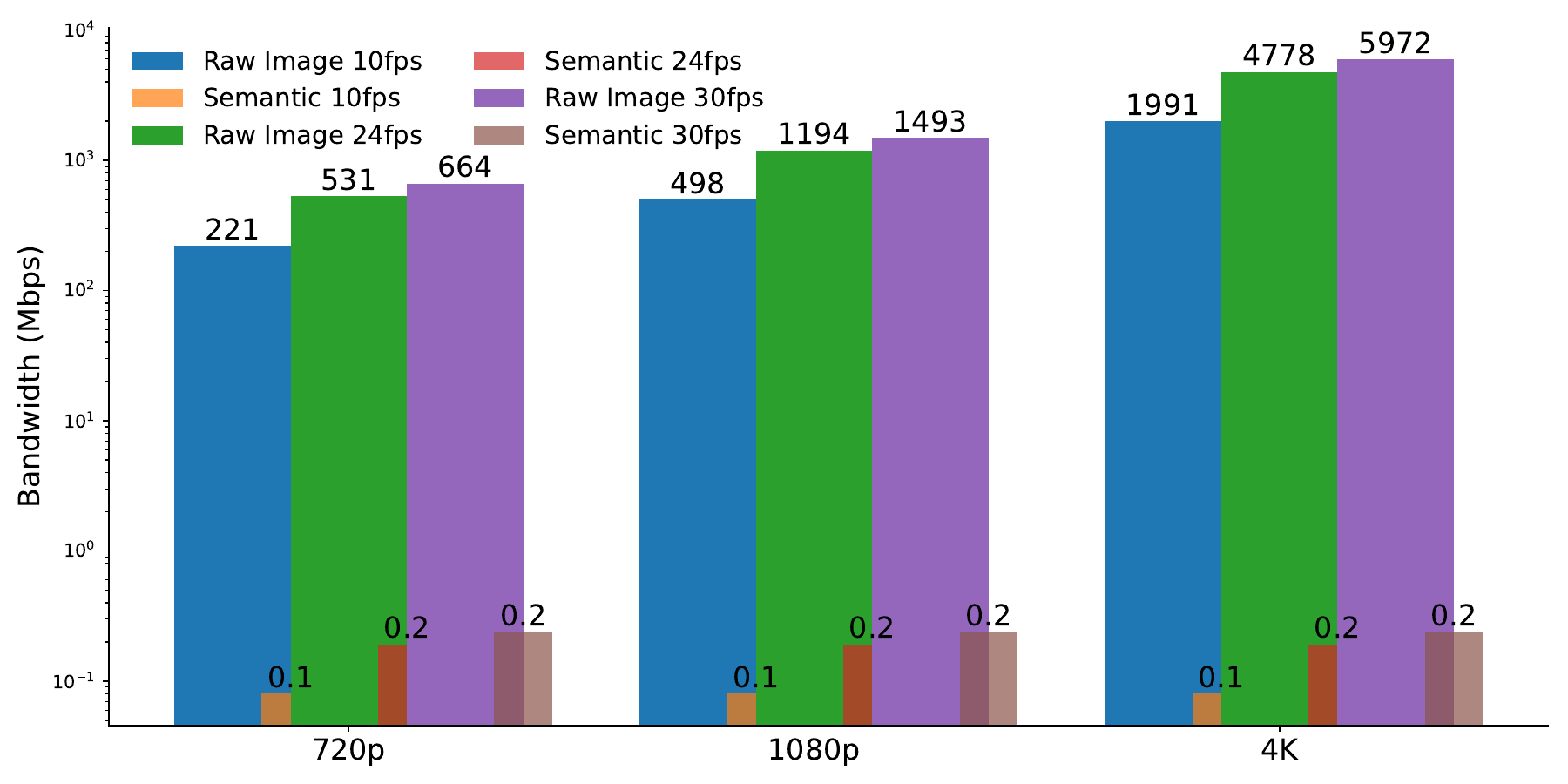}
\caption{Bandwidth requirement comparison between traditional cloud-based image transmission and semantic communication with onboard processing. The semantic approach, enabled by edge AI processing, transmits only essential structured data, achieving significant bandwidth efficiency across different video resolutions and frame rates.}
\label{fig:bandwidth}
\end{figure}

We implemented a road quality classification and a pothole detection model on the Nvidia Jetson Orin Nano device as part of the road condition monitoring task. The classification model is based on EdgeFusionViT \cite{10510402} and the segmentation model utilizes YOLOv8\footnote{\url{github.com/ultralytics/ultralytics}}. In our preliminary tests, the edge device required approximately $80$ ms to classify road quality and about $115$ ms to detect potholes per frame. However, these times do not account for the data transmission between the edge device, other drones in the swarm, and the LLM servers. \figureautorefname~\ref{fig:jetson_output} demonstrates the performance of the embedded device. The classification model identifies the material, friction level, and unevenness level of the road, and the segmentation model identifies the location of potholes. In our preliminary experiments, we have observed satisfactory performance from such a resource-constrained device. Our initial experiments also indicate that covering a moderately large geographical area in real-time with a swarm of drones necessitates the adoption of 6G-enabled technologies to ensure seamless and efficient operations.
The efficiency of our onboard processing approach compared to traditional cloud-based systems is demonstrated in \figureautorefname~\ref{fig:bandwidth}, which compares data transmission requirements between conventional image-based systems and our SC framework. The figure illustrates bandwidth requirements across different video resolutions and frame rates, showing that SC, which transmits structured LLM-generated data containing essential attributes,  requires significantly lower bandwidth compared to raw image transmission.

\subsection{Network Performance Analysis}
To quantitatively evaluate the performance differences between 5G and 6G networks in drone swarm coordination, we developed a simplified mathematical model for collision probability and fault detection latency. The evaluation incorporates network parameters based on existing 5G specifications (latency: $1$ ms, reliability: $99.999\%$) \cite{baek20213gpp} and projected next-generation 6G requirements (latency range: $0.1-1$ ms with $0.5$ ms as representative value, reliability: $99.999999\%$) \cite{shamsabadi2025exploring6gpotentialsimmersive}. The collision probability $P_c$ for a swarm of $n$ drones is modeled as: $P_c(n) = \alpha_k \cdot \frac{n}{n_0} \cdot (1 - R_k)$. Here, $\alpha_k$ represents the base collision rate ($\alpha_{5G}=0.02$, $\alpha_{6G}=0.001$), $n_0$ is the reference swarm size, and $R_k$ denotes the network reliability. For fault detection time $T_d$, we model both the network latency and congestion effects: $T_d = L_b \cdot (1 + \frac{n}{n_{max}}) + \epsilon$. Here, $L_b$ represents the base network latency, $n_{max}$ is the maximum supported swarm size, and $\epsilon$ follows an exponential distribution with mean $L_b$. Table \ref{tab:network_performance} presents our simulation results comparing 5G and 6G performance across different swarm sizes. The results include collision rates (CR) and detection times (DT) with their standard deviations and $95\%$ confidence intervals (CI), quantitatively demonstrating how 6G's enhanced capabilities translate to superior performance in both metrics. The simulation implements a Monte Carlo approach with $100$ iterations per configuration. For each iteration, fault occurrences follow a Poisson distribution with mean, $\mu=5$ faults per simulation period. $\mathcal{N_D}$ represents the number of drones.

\begin{table}[htbp]
\scriptsize
\centering
\caption{Network Performance Analysis}
\label{tab:network_performance}
\begin{tabular}{cccccc}
\toprule
$\mathcal{N_D}$ & Net & CR (\%) & DT (ms) & CR 95\% CI & DT 95\% CI \\
\midrule
10 & 5G & 1.995 ± 0.052 & 2.12 ± 0.41 & [1.912, 2.099] & [1.49, 3.05] \\
10 & 6G & 0.101 ± 0.013 & 1.10 ± 0.23 & [0.077, 0.128] & [0.73, 1.58] \\
20 & 5G & 4.003 ± 0.072 & 2.34 ± 0.43 & [3.877, 4.133] & [1.58, 3.35] \\
20 & 6G & 0.200 ± 0.020 & 1.21 ± 0.26 & [0.165, 0.234] & [0.81, 1.78] \\
30 & 5G & 6.013 ± 0.093 & 2.70 ± 0.60 & [5.820, 6.165] & [1.86, 4.04] \\
30 & 6G & 0.303 ± 0.022 & 1.26 ± 0.20 & [0.260, 0.343] & [0.92, 1.66] \\
40 & 5G & 7.999 ± 0.100 & 2.77 ± 0.42 & [7.786, 8.159] & [2.07, 3.63] \\
40 & 6G & 0.398 ± 0.024 & 1.39 ± 0.26 & [0.356, 0.445] & [1.03, 2.03] \\
50 & 5G & 10.011 ± 0.125 & 2.96 ± 0.59 & [9.778, 10.256] & [2.14, 3.87] \\
50 & 6G & 0.503 ± 0.028 & 1.54 ± 0.28 & [0.450, 0.555] & [1.19, 2.22] \\
\bottomrule
\end{tabular}
\end{table}

\section{Conclusion}
In this research, we have proposed a 6G-enabled drone swarm system for autonomous infrastructure monitoring by integrating URLLC, edge AI, SC, and LLMs. This approach addresses the limitations of current 5G networks in handling the real-time coordination and decision-making required for large-scale automated inspection systems. Our proposed framework utilizes 6G technologies to perform complex inspection tasks and generate reports from semantically processed multimodal data. We have also identified the critical drivers and enabling technologies of 6G necessary for the success of such systems.

Future research should focus on refining the proposed system through real-world deployments and simulations. We also envision that 6G will drive the proliferation of digital twins, allowing virtual interactions with physical infrastructures. This integration could significantly aid in the early prediction of faults or errors and make infrastructure management more interactive and accessible, even for non-experts. By utilizing digital twins, the system could offer a more intuitive interface for users, allowing for more proactive and effective infrastructure management.

\bibliographystyle{IEEEtran}
\bibliography{ref}

\vfill

\end{document}